\documentclass[conference]{IEEEtran}
\IEEEoverridecommandlockouts
\usepackage{graphicx}
\usepackage{array}
\usepackage{longtable}
\usepackage{color}
\usepackage{amsmath}
\usepackage{amsfonts}
\usepackage{amssymb}
\usepackage{graphics}
\usepackage{slashbox}

\usepackage[margin=2cm]{geometry}
\usepackage{adjustbox}
\usepackage{multirow}

\usepackage[utf8]{inputenc}
\usepackage{graphicx}
\usepackage{pgfplots}
\usepackage{makecell}
\usepackage{caption}
\usepackage{subcaption}

\usepackage{mwe}
\usepackage{ragged2e}

\pgfplotsset{compat=1.14}

\newcommand{\be}{\begin{eqnarray}}
\newcommand{\ee}{\end{eqnarray}}
\newcommand{\bee}{\begin{eqnarray*}}
\newcommand{\eee}{\end{eqnarray*}}

\newcommand{\matrixb}{\left[ \begin{array}}
\newcommand{\matrixe}{\end{array} \right]}

\begin{document}

\title{Sequential Deep Learning Architectures for Anomaly Detection in Virtual Network Function Chains
\thanks{}
}

\author{\IEEEauthorblockN{Chungjun Lee\IEEEauthorrefmark{1},
Jibum Hong\IEEEauthorrefmark{2}, DongNyeong Heo\IEEEauthorrefmark{1}, Heeyoul Choi\IEEEauthorrefmark{1}}
\IEEEauthorblockA{
\IEEEauthorrefmark{1}Department of Information Communication Engineering Handong Global University, Pohang, South Korea\\
\{chunglee3224, sjglsks, heeyoul\}@gmail.com\\
\IEEEauthorrefmark{2}Department of Computer Science and Engineering, POSTECH, Pohang, South Korea\\
\IEEEauthorrefmark{2}hosewq@postech.ac.kr}}

\maketitle
\begin{abstract}
Software-defined networking (SDN) and network function virtualization (NFV) have enabled the efficient provision of network service. However, they also raised new tasks to monitor and ensure the status of virtualized service, and anomaly detection is one of such tasks.
There have been many data-driven approaches to implement anomaly detection system (ADS) for virtual network functions in service function chains (SFCs). In this paper, we aim to develop more advanced deep learning models for ADS.
Previous approaches used learning algorithms such as random forest (RF), gradient boosting machine (GBM), or deep neural networks (DNNs). However, these models have not utilized sequential dependencies in the data.
Furthermore, they are limited as they can only apply to the SFC setting from which they were trained. 
Therefore, we propose several sequential deep learning models to learn time-series patterns and sequential patterns of the virtual network functions (VNFs) in the chain with variable lengths.
As a result, the suggested models improve detection performance and apply to SFCs with varying numbers of VNFs.
\end{abstract}


\section{Introduction}
Softwarization of computer networks has enabled efficient ways to deploy and manage services with flexibility and reconfigurability. For example, hardware-based network functions can now be replaced with virtual network functions to release constraints related to hardware. In addition, cloud computing with software-based networks enables service providers to utilize computing resources for service management in data centers more efficiently \cite{Hong2020}.

Meanwhile, the number of network devices and linkages has grown over the years \cite{Nedelkoski2019}, and the softwarization of computer networks also has increased the number of services provided through virtual networks. The increased complexity and size of the network introduced problems of quality degradation and inconsistency in service availability. This extension in the network size poses a severe management challenge \cite{Hong2020, Nedelkoski2019}.

As service providers are aware that they need to provide dependable service, they are interested in a data-driven approach for virtual network management, combining big data and machine learning (ML) to manage large-scale networks with complex dependencies efficiently. Among much prior work in this line of research \cite{network_Heo2020, network_Lange2019, network_Gwon2019}, one of the vital network management techniques is anomaly detection, in which the quality of service is preserved by quickly responding to a system that shows out-of-normal behavior. 

Related to ML-based anomaly detection for virtual network functions (VNFs), previous works trained ML models with data collected from an operating system (OS) or hyper-visor of virtual machines (VMs) \cite{Hong2020, Sauvanaud2016, Sauvanaud2018}. For example, in \cite{Hong2020}, they collected a dataset and applied methods such as decision tree-based gradient boosting machine (GBM), XGBoost, and deep neural networks (DNNs) to detect anomalous cases from the normal ones. 

\begin{figure}[h]
\centerline{\hbox{ 
\includegraphics[width=3.4in]{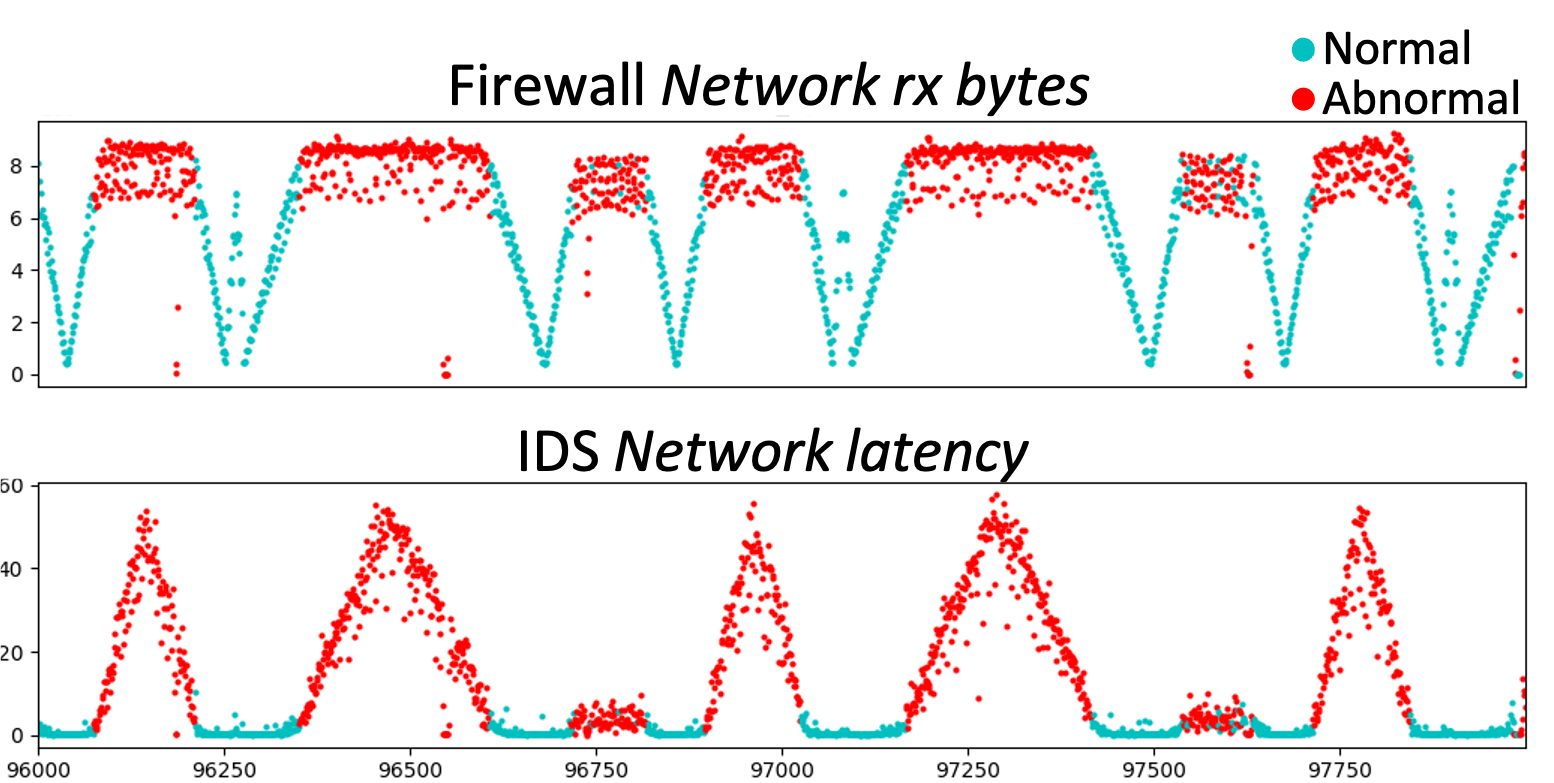}
}}
\caption{Time-series plot of anomaly detection data. The figure shows a time-series plot of selected features from an anomaly detection dataset. The plot shows a sequential pattern related to the occurrence of abnormality, where the x and y axes indicate the time index and feature values at each time step, respectively. Cyan and red are for normality and abnormality, respectively.}
\label{fig:time_series_plot}
\end{figure}

However, these methods have not utilized sequential information over the VNF sequence, which should be helpful for anomaly detection as VNF instances are arranged in sequence over which network traffic passes \cite{Cotroneo2017}. In addition, if services require different compositions of SFCs, including type, order, and the number of VNF instances, existing approaches had to train separate models, especially when the number of VNF instances is different because of the discrepancy in input dimension. Furthermore, they have not employed temporal information over time-series of monitoring data, which should be helpful if occurrences of abnormality follow temporal patterns as in Figure \ref{fig:time_series_plot}. 

To learn sequential information over the VNF sequence, we apply uni and bi-directional recurrent neural networks (Uni- and Bi-RNNs) and Transformer.  As the sequential architectures can take a variable-length input sequence, our proposed models are compatible with data from SFCs with different numbers of VNFs. Such compatibility allows the capacity of joint training (or training simultaneously) over data from different SFCs. Moreover, to learn time-series information of monitoring data, we use RNNs to take the input of the monitoring data at a given time-step and a series of inputs from previous time-steps.

In order to increase the benefit of learning time-series information from monitoring data, we add another component to the proposed models to utilize the predicted result from the preceding time step. This is the same mechanism as in natural language processing (NLP) models like language model and neural machine translation \cite{Bengio2015, readout_Bahdanau2015}. These modified models use prediction of the preceding time-step as additional input to the current time step in RNNs, which takes advantage of temporal patterns while maintaining the benefits of the initially proposed models.

The proposed models improved detection performance over baseline models in three anomaly detection datasets in our experiment result. Further, the results show that joint training gives a similar level of performance to that of the models trained on each dataset individually. Also, in our experiment setting, the modified models, which use the preceding prediction result, provide additional improvement for both individual and joint training cases.

The rest of the paper is organized as follows: Section 2 describes the background for the proposed model. In section 3, we explain the motive and the proposed deep learning models. In section 4, the experiment process and results are presented, followed by a conclusion in section 5.

\section{Background}
\subsection{Anomaly Detection Methods}
Anomaly detection is a binary classification task where the model determines whether a given system is in anomalous status. There are two types of anomalies, including network anomaly and system anomaly which are related to network traffic pattern and system resource usage. This paper focuses on both types of anomalies, which lead to degradation of service quality marked by service level violation (SLA).

In ML-based anomaly detection in the virtual network management domain, prior literature simulated testbed environments that generate normal and abnormal patterns of monitoring data from each component where abnormal data is generated by anomaly injection techniques, which causes a surge effect in OS resource usage or network traffic \cite{Sauvanaud2016, Sauvanaud2018, Hong2020}. One previous research simulated IP multimedia service (IMS) environment and anomaly injection techniques to collect and label data, and it utilized various ML algorithms such as random forest (RF) to show strong detection performance \cite{Sauvanaud2016}. Another prior literature simulated a network environment that utilizes network function virtualization (NFV), where VNFs were deployed and connected through SFC in web service and login authentication scenarios, and an array of ML algorithms were trained and evaluated over the collected dataset \cite{Hong2020}. In this paper, we follow the experiment setting of \cite{Hong2020} for anomaly detection in SFC based on monitoring data from the VNFs. 


There is much recent work that uses an unsupervised learning model. In contrast to supervised learning approaches, they do not require labeling data because they are trained on only normal patterns of monitoring data \cite{Gulenko2018, Schmidt2018arima, Schmidt2018iftm}. Such models can measure the deviance of observed monitoring data at test time and raise an alert when the deviance is significant. One of the approaches computes a set of centroids using the BIRCH algorithm to define the boundaries for normal and abnormal data \cite{Gulenko2018}. Other approaches utilize various algorithms for modeling time-series forecasting such as autoregressive integrated moving average (ARIMA) or recurrent neural networks (RNNs) to learn patterns of monitored data at normal situations for anomaly detection \cite{Schmidt2018arima, Schmidt2018iftm}. Unsupervised learning method has benefits compared to supervised learning in that it is applicable even if labeling data is not available, while unsupervised learning models lacked performance compared to supervised learning models \cite{Lee2020}.

\subsection{Sequential Deep Learning Models}
Sequential deep learning models include Uni(or Bi)-RNNs and Transformer. RNNs have a recurrent connection to capture sequential patterns in data, and Bi-RNNs have stacked forms of RNNs to read the sequence in both directions \cite{rnn_Hochreiter1997, rnn_Schuster1997}. Transformers are a neural network architecture that can compute the representation of sequence data based on self-attention without explicit recurrent connections \cite{transformer_vaswani2017}. Transformer networks showed state-of-the-art performance in modeling sequential patterns, especially in NLP. More detailed descriptions of sequential deep learning models can be found in \cite{rnn_Hochreiter1997, transformer_vaswani2017}.

\begin{figure}[h]
\centerline{\hbox{
\includegraphics[width=3.0in]{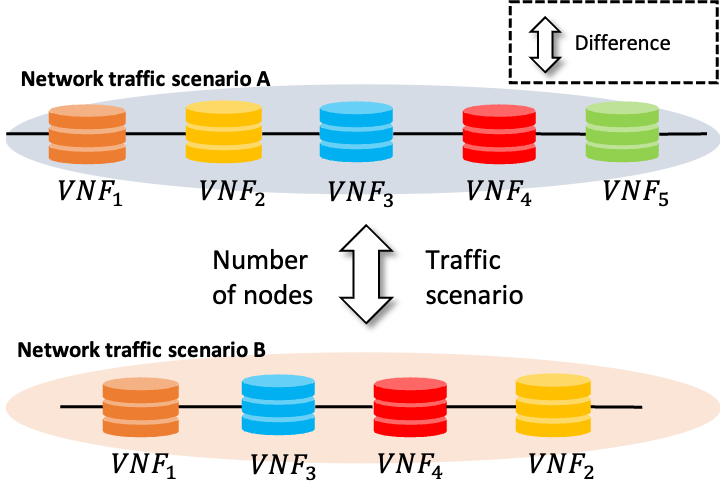}
}}
\caption{SFCs with varying numbers of VNFs. The figure describes two different SFCs deployed in different traffic environments. Our proposed models have compatibility with monitoring data from different numbers of VNFs. Some of the shapes in this figure are adopted from \cite{network_Heo2020}.}
\label{fig:variable-length-sfcs}
\end{figure}

\section{Proposed Method}
\subsection{Problem definition}
In the previous approach \cite{Hong2020}, the problem can be defined as learning a parametric function $\hat{y}_t = f(X_{t}; \theta)$ which produces a binary output $\hat{y}_t$ where $t$ and $\theta$ are time index and parameter of the model, respectively. Also, $X_t \in \mathbb{R}^{V \times D_{input}}$ is the input, where $V$ is number of VNFs in the SFC, and $D_{input}$ is the number of selected metrics (or features) from a VNF instance such as CPU or memory utilization rates.

In our approach, we consider both the input at current time index and previous series of inputs. Therefore, input is $X_{t-l+1}^{t} = [X_{t-l+1}, X_{t-l+2}, ..., X_{t}]$, where $l$ is the sequence length. Additionally, each $X_i$ is considered as a set of monitored data from each VNF instance, $X_i = [x_{i, 1}, x_{i, 2}, ... x_{i, V}] $ where $x_{i,v} \in \mathbb{R}^{D_{input}}$ with $t \in [l,...,T]$. Here, $T$ and $i$ denote respectively the number of samples and arbitrary time index. Note that $X_i$ is a sequence of several VNFs and the models should be compatible with different sizes. 

\subsection{Proposed Neural Network model}
The proposed model is composed of feature mapping, encoder, readout, and classifier layers. The first three layers enable learning sequential patterns over VNF sequence with variable lengths where input $X_i$ of arbitrary sequence length is encoded into fixed-length vector $\tilde{z}_i$. Therefore, proposed architectures improve performance and are compatible with monitoring data collected from different SFCs, as shown in Figure \ref{fig:variable-length-sfcs}.

First, the feature mapping layer is fully connected layer without activation function, and it maps input vector $x_{i,v}$ to $\tilde{x}_{i,v} \in \mathbb{R}^{D_z}$, where $D_z$ is the dimension of feature mapping. Second, the encoder layer is sequential deep learning architecture like Uni-RNN, Bi-RNN and Transformer. It takes input sequence $\tilde{X}_i = [\tilde{x}_{i,1}, \tilde{x}_{i,2}, ..., \tilde{x}_{i,V}]$ and outputs $Z_i = [z_{i,1}, z_{i,2}, ..., z_{i,V}]$. Third, the readout layer is one of functions like \emph{max}, \emph{mean}, or \emph{self-attention}. \emph{Max} and \emph{mean} are conventional pooling methods to combine multiple and varying numbers of input vectors. The \emph{self-attention} mechanism summarizes the set of vectors by a weighted sum, where the weights can be computed based on the sequence with parameters, $\theta_{att}$ \cite{readout_Bahdanau2015, hchoi2018neuro}. Output of the readout layer is $\tilde{z}_i = readout([z_{i,1}, z_{i,2}, ..., z_{i,V}])$, where $\tilde{z}_i \in \mathbb{R}^{D_z}$. Note the first three layers apply to monitoring data from SFC with different numbers of VNFs. The forward equation of the first three layers are summarized as,
\begin{eqnarray}
\tilde{X}_i &=& \text{\emph{feature\_mapping}}(X_i; \theta_{map}),\\
Z_i &=& \text{\emph{encoder}}(\tilde{X}_i; \theta_{enc}),\\
\tilde{z}_i &=& \text{\emph{readout}}(Z_i; \theta_{att}).
\end{eqnarray}

The classifier layer is composed of RNN and DNN. This layer captures sequential patterns existing over time-steps. In this layer, RNN takes input sequence, $[\hat{z}_{t-l+1}, \hat{z}_{t-l+2}, ..., \hat{z}_t]$, and outputs $[h_{t-l+1}, h_{t-l+2}, ..., h_{t}]$. Then DNN takes RNN hidden state, $h_t$ as input to produce the detection result, $\hat{y_t} = DNN(h_t)$. In summary, the forward equations of the classifier layer and the objective function $J$ for the binary classification are given as,
\begin{eqnarray}
h_{t-l+1}^{t} &=& \text{\emph{RNN}}(\tilde{z}_{t-l+1}^{t}; \theta_{RNN}),\\
\hat{y}_t &=& \text{\emph{DNN}}(h_t; \theta_{DNN}),\\
J(y_t, \hat{y_t}) &=& -y_tlog(\hat{y_t})-(1-y_t)log(1-\hat{y}_t).
\end{eqnarray}

Here, the trainable parameter includes $\theta_{map}, \theta_{enc}, \theta_{att}, \theta_{RNN}$, and $\theta_{DNN}$. They are optimized using gradient-descent method to minimize the objective function $J$ which is the binary cross-entropy (BCE) \cite{Graves2012}.

\begin{figure}[h]
\centerline{\hbox{ 
\includegraphics[width=3.0in]{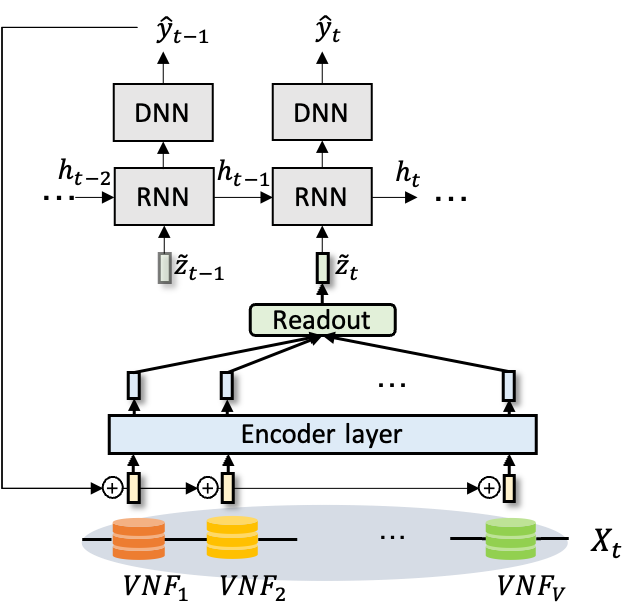}
}}
\caption{Overview of the proposed model. The feature mapping, encoder, and readout layers process input sequence into fixed dimensional vector, $\tilde{z}_t$. At each time-step, input data, $X_t$ is concatenated with prediction result from previous time-step, $\hat{y}_{t-1}$., so that $[X_{t-l+1}, X_{t-l}, ..., X_{t}]$ is forwarded to $[\tilde{z}_{t-l+1}, \tilde{z}_{t-l}, ..., \tilde{z}_t]$ and to the classifier layer. Note the feature mapping layer is omitted for brevity.}
\label{fig:model_architecture}
\end{figure}

Additionally, we use prediction result from the previous time-step for further improvement. As we observed that incidents of abnormality are highly correlated to the event in the preceding data point, inputting prediction results of preceding time-step should be beneficial to prediction at the current time-step. To use this information, we modify the model to use previous prediction result, $\hat{y}_{i-1}$ together with current input $X_i$ by concatenation. Therefore, the input is changed to $X_i = [(\hat{y}_{i-1} \oplus x_{i,1}), (\hat{y}_{i-1} \oplus x_{i,2}), ..., (\hat{y}_{i-1} \oplus x_{i,V})]$. The overview of our proposed model is shown in Figure \ref{fig:model_architecture}.

\section{Experiment}
This section provides a brief description of the experiment setting, data acquisition and presents results with analysis. We designed experiments to check the benefits of two contributions in the proposed models: improvement of detection performance and compatibility.

\subsection{Virtual Network Setup}
We set up two virtual network scenarios from which we collect anomaly detection data. In the web hosting service scenario, the client sent requests to the webserver, going through 5 VNFs: FW, IDS, FM, DPI, and LB, and we refer to the dataset collected in this setting as web service data (WSD). In the login authentication scenario, the server was connected to the database system. The client sent requests to the server through 4 VNFs: FW, FM, DPI, and IDS, which leads to login authentication data (LAD). Depending on how strict the SLA was, two sets were gathered: LAD1 and LAD2 (See the next section). The topology of the testbed \cite{dpnm} was closely related to the multi-access edge computing (MEC) scenario, and OpenStack (rocky release) was used for constructing virtual network and SFC. Open source VNFs were considered, such as firewall (FW, iptables \cite{firewall}), intrusion detection system (IDS, Suricata \cite{ids}), flow monitor (FM, ntopng \cite{flow_monitor}), deep packet inspection (DPI, nDPI \cite{dpi_Deri2014}), and load balancer (LB, HAProxy \cite{lb}). 

\subsection{Data Acquisition Components}
To gather data, our ADS had components for monitoring, fault injection, metric selection, and data labeling. A more detailed description of each component of ADS can be found in \cite{Hong2020}.
\subsubsection{Monitoring}
From VMs which composed virtual network, we monitored metrics related to CPU, memory, disk I/O, and network traffic. The monitoring function was composed of a monitoring agent, service, and dashboard. Collectd \cite{collectd}, InfluxDB \cite{influxDB}, and Grafana \cite{grafana} were used for the implementation of each component.

\subsubsection{Fault Injection}
Since anomalies do not exist frequently in a given virtual network setting, various software and hardware faults were injected into the system on which VNFs were deployed. The first fault injection method was a generation of abnormal states to VMs: CPU utilization, memory usage, disk I/O, network latency, and network packet loss through a tool `stress-ng' \cite{stress-ng}. The second method was a generation of heavy workload by sending tremendous network traffic using D-ITG traffic generator \cite{Botta2012}.

\subsubsection{Metric Selection and Data Labeling}
We selected input metrics from monitoring as shown in Table \ref{tbl:metrics}. The labeling of the data was based on SLA-related metrics as in \cite{GRAAP}. As the response time and availability were measured, the monitoring data collected when response time (less than 250ms) or availability (over 99.95\% success rate of requests) was not satisfied are labeled as an anomaly. The LAD2 dataset applies a more strict SLA than that for WSD and LAD1 datasets (less than 200ms for response time and 99.99 \% success rate of requests). Statistics of the datasets are summarized in Table \ref{tbl:dataset_stat}.

\begin{table}[h!] \centering
\caption{Selected metrics for anomaly detection, adopted from \cite{Hong2020}}
    \label{tbl:metrics}
    \begin{tabular}{| c | l | c | l |}
    \hline
    \textbf{Metrics} & \textbf{Descriptions} \\ 
    \hline \hline
   time & Measurement time \\
   instance & VNF instance name \\
   cpu\_idle & CPU - idle time \\
   cpu\_interrupt & CPU - interrupt time \\
   cpu\_nice & CPU - nice status time \\
   cpu\_softirq & CPU - softirq time \\
   cpu\_steal & CPU - stolen time \\
   cpu\_system & CPU - used by kernel mode \\
   cpu\_user & CPU - used by user mode \\
   cpu\_wait & CPU - I/O wait time \\
   mem\_free & Memory - free space \\
   mem\_buffered & Memory - buffered space \\
   mem\_cached & Memory - cached space \\
   mem\_used & Memory - used space \\
   disk\_free & Disk - free space \\
   reserved & Disk - reserved space \\
   disk\_used & Disk - used space \\
   io\_read & I/O read bytes \\
   io\_write & I/O write bytes \\
   io\_time & I/O - spent time \\
   network\_rx\_bytes & Received traffic bandwidth \\
   network\_tx\_bytes & Transmitted traffic bandwidth \\
   network\_rx\_packets & The number of received packets \\
   network\_tx\_packets & The number of transmitted packets \\
   network\_latency & Hop latency between VNFs \\
   \hline
   \end{tabular}
\end{table}

\begin{table}[h!] \centering	
\caption{Statistics of anomaly detection datasets. The numbers of anomalies for training, validation, and test sets are in the parentheses.}
    \label{tbl:dataset_stat}
    \begin{tabular}{| l | r | r | r |}
    \hline
    \textbf{Dataset}  & \textbf{ WSD } & \textbf{ LAD1 } & \textbf{ LAD2 }\\ 
    \hline \hline
   VNF instances & 5 & 4 & 4 \\
   \hline
   Total samples  & 68,731 & 121,053 & 121,024 \\
   Anomalies  & 26,354 & 19,913 & 44,513 \\
   \hline
   Training set & 44,665 (16,933) & 78,675 (12,910) & 78,656 (28,734) \\
   Validation set & 6,872 (2,690) & 12,103 (1,997) & 12,100 (4,465) \\
   Testing set & 17,178 (6,731) & 30,259 (5,006) & 30,252 (11,314) \\
   \hline
   \end{tabular}
\end{table}

\begin{table*}[h!]
  \centering
  \caption{Performances of initial models which do not employ prediction result of the preceding time-step. }
  \label{tbl:improvement_in_detection_performance}
  \begin{tabular}{|c|cc|c|c|c|}
    \hline
      \multirow{2}{*}{\shortstack[*]{\textbf{Training setting}}} & \multicolumn{2}{|c|}{\bfseries Model} & \bfseries WSD & \bfseries LAD1 & \bfseries LAD2 \\
    \cline{2-6}
      & \bfseries Encoder & \bfseries Readout & \bfseries F1-measures & \bfseries F1-measures & \bfseries F1-measures \\
    \hline
    \multirow{6}{*}{\shortstack[*]{\textbf{Individual training}}} & \multicolumn{2}{|c|}{GBM [1]} & 96.19 & 95.20 & 93.27 \\
    & \multicolumn{2}{|c|}{XGBoost [1]} & 95.81 & 95.30 & 93.81 \\
    & \multicolumn{2}{|c|}{DNN [1]} & 93.14 & 90.27 & 90.77 \\
    \cline{2-6}
    & RNN & \multirow{3}{*}{\shortstack[*]{Max / Mean /\\ Self-attention}} & 97.52 / 97.46 / 97.65 & 97.64 / 97.77 / 97.75 & 95.79 / 96.01 / 95.77 \\
    & Bi-RNN & & 97.59 / 97.72 / 97.59 & 97.95 / 97.95 / 97.77 & 95.74 / 95.97 / 95.81 \\
    & Transformer & & 97.69 / 98.11 / 98.04 & 97.69 / 97.87 / 97.83 & 95.79 / 95.88 / 96.00 \\\hline
    \multirow{3}{*}{\bfseries WSD \& LAD2} & RNN & \multirow{3}{*}{\shortstack[c]{Max / Mean /\\Self-attention}} & 97.49 / 97.52 / 97.74 & & 94.91 / 95.04 / 95.19 \\
    & Bi-RNN & & 97.43 / 97.50 / 97.68 & NA & 95.09 / 95.11 / 95.01 \\
    & Transformer & & 97.91 / 98.17 / 98.04 & & 95.64 / 95.30 / 95.45 \\\hline
  \end{tabular}
\end{table*}

\begin{table*}[h!]
  \centering
  \caption{Performances of the modified models which utilize the output of the previous time-step.}
  \label{tbl:improvement_in_detection_performance_modified}
  \begin{tabular}{|c|cc|c|c|c|}
    \hline
      \multirow{2}{*}{\shortstack[*]{\textbf{Training setting}}} & \multicolumn{2}{|c|}{\bfseries Model} & \bfseries WSD & \bfseries LAD1 & \bfseries LAD2 \\
    \cline{2-6}
      & \bfseries Encoder & \bfseries Readout & \bfseries F1-measures & \bfseries F1-measures & \bfseries F1-measures \\
    \hline
    \multirow{3}{*}{\shortstack[*]{\textbf{Individual training}}} & RNN & \multirow{3}{*}{\shortstack[*]{Max / Mean /\\ Self-attention}} & 100.00 / 100.00 / 100.00 & 99.98 / 99.99 / 99.98 & 99.97 / 99.94 / 99.96 \\
    & Bi-RNN & & 99.99 / 99.99 / 99.99 & 99.99 / 99.99 / 100.00 & 99.95 / 99.97 / 99.97 \\
    & Transformer & & 99.99 / 100.00 / 100.00 & 99.96 / 99.98 / 99.98 & 99.97 / 99.95 / 99.98 \\\hline
    \multirow{3}{*}{\bfseries WSD \& LAD2} & RNN & \multirow{3}{*}{\shortstack[c]{Max / Mean /\\Self-attention}} & 99.99 / 99.98 / 99.99 & & 99.92 / 99.91 / 99.90 \\
    & Bi-RNN & & 99.99 / 99.99 / 99.98 & NA & 99.94 / 99.92 / 99.89 \\
    & Transformer & & 100.00 / 100.00 / 100.00 & & 99.94 / 99.91 / 99.92 \\\hline
  \end{tabular}
\end{table*}

\subsection{Experiment Result}
We trained and tested our models on three different datasets: WSD, LAD1, and LAD2. 

\subsubsection{Initial Models}
First, we test our models without the detection result of the previous time-step, $\hat{y}_{t-1}$.
The detection performance of the proposed models are presented in Table \ref{tbl:improvement_in_detection_performance}. In the table, The first group of models is baselines where GBM and XGBoost are best performing ML models from prior literature \cite{Hong2020}. For the second group of models, we experimented with possible combinations of an encoder (RNN, Bi-RNN, and Transformer) and a readout method (Max, Mean, and self-attention). For example, F1-measures for Bi-RNN + Max and Transformer + Mean on LAD1 are 97.95 and 97.87, respectively. 

The training setting, `individual training,' shows that one model is trained and evaluated in each dataset. On the other hand, `WSD \& LAD2' indicate that one model was jointly trained on both datasets and tested on each of them. Note that LAD1 and LAD2 share the same environment from which the dataset was generated, but the difference is labeling criteria where LAD2 applied more strict and suitable criteria for the given setting \cite{Hong2020}. Therefore, we carried out a joint training experiment only with `WSD \& LAD2' datasets.

As shown in the table, the proposed models improve F1-measure over the baselines. In addition, the models have an additional advantage in that they can use monitoring data from variable length SFCs as input, enabling joint training. In given datasets, jointly trained models show a similar level of performance as those trained exclusively on the dataset. Such a result demonstrates the potential of the proposed models that they can be used for different SFC setups without requiring two separate detection models.

\begin{figure}
\captionsetup[subfigure]{justification=Centering}

\begin{subfigure}[t]{\linewidth}
    \includegraphics[width=3.4in]{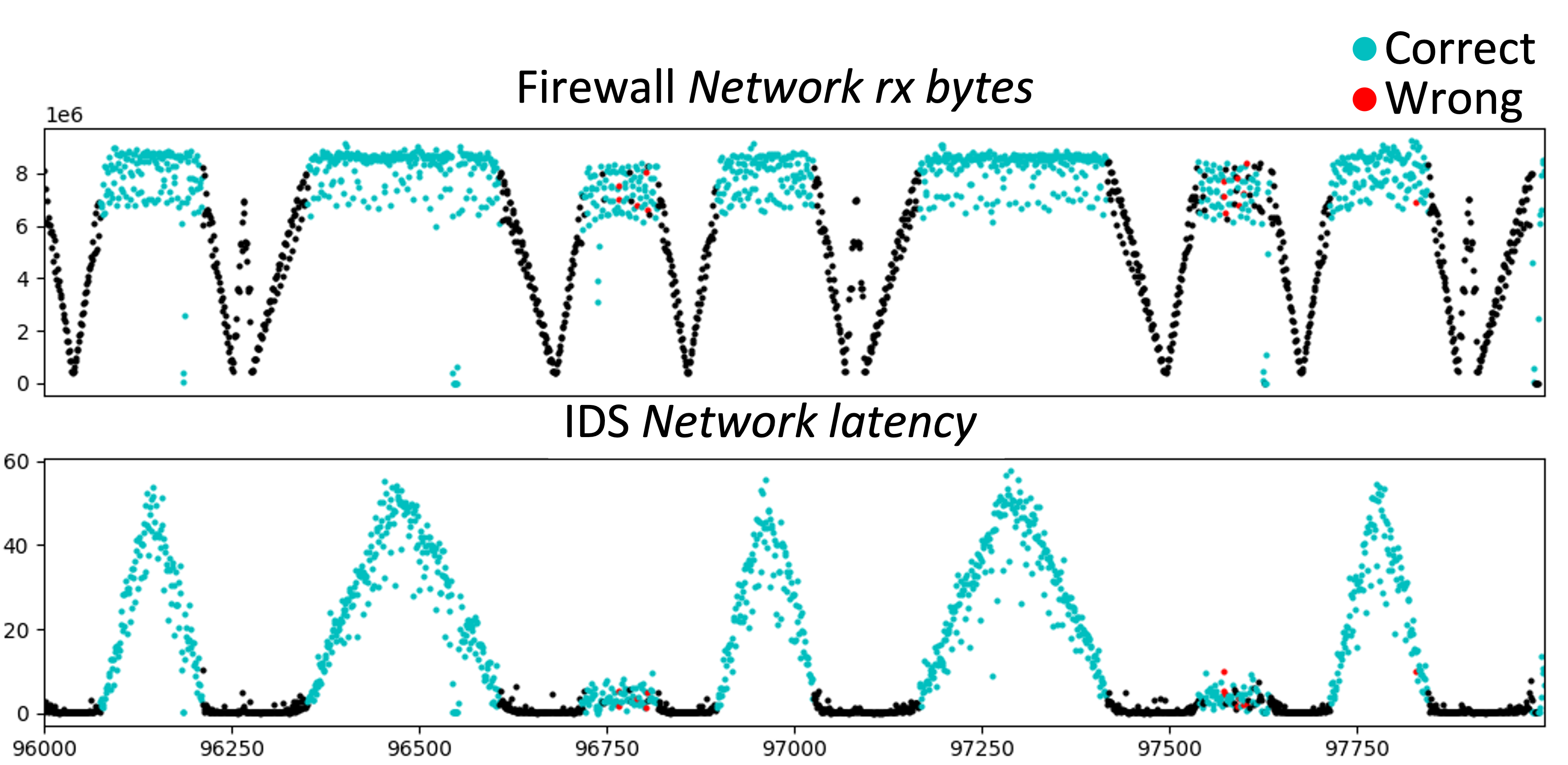}
    \caption{Detection result of Transformer + self-attention, which does not utilize the detection results in the previous time-step.}
\end{subfigure}
\newline
\begin{subfigure}[t]{\linewidth}
    \includegraphics[width=3.4in]{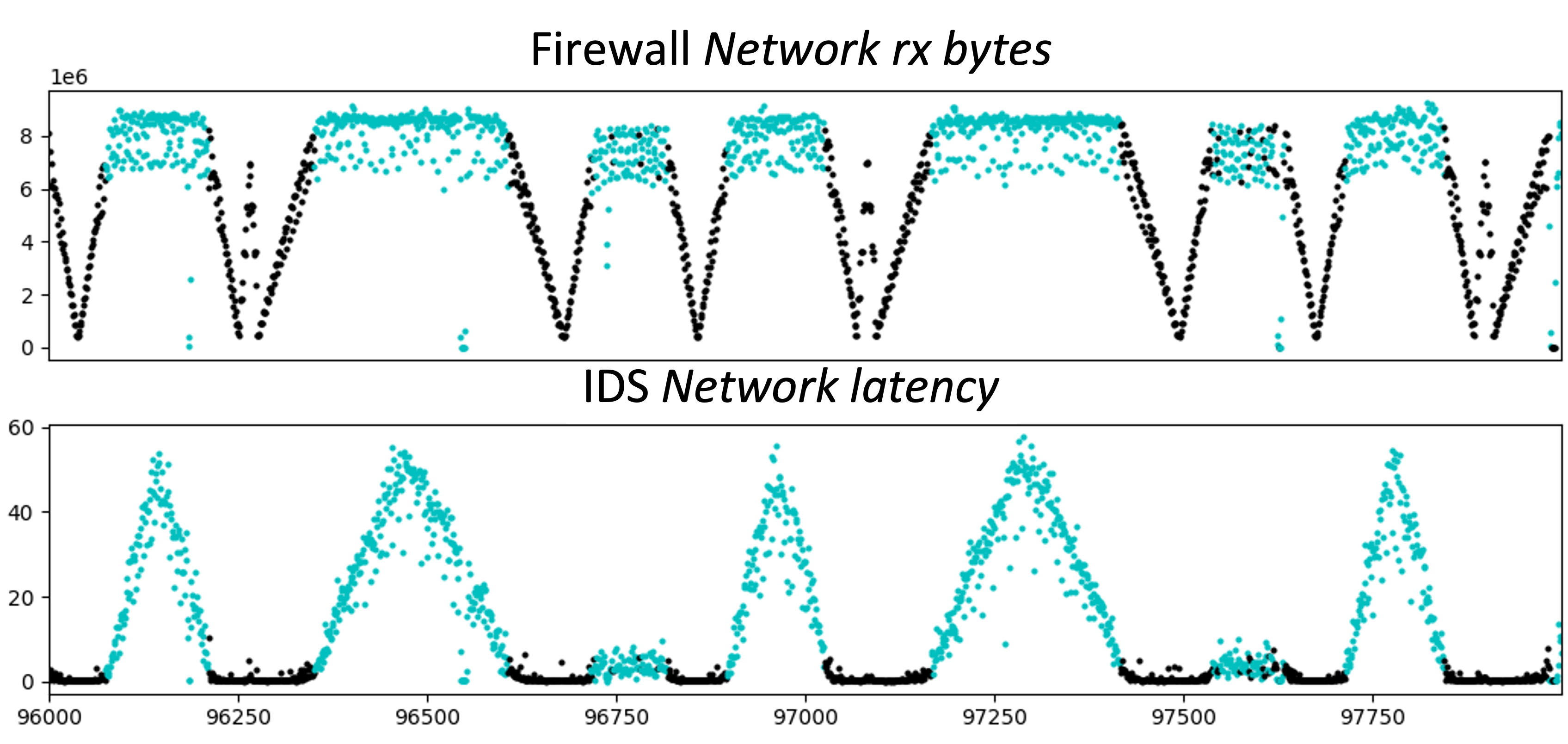}
    \caption{Detection result of Transformer + self-attention, which utilizes the detection results in the previous time-step, $\hat{y}_{t-1}$}.
\end{subfigure}
\caption{Performance comparison between models with DNN classifier and RNN classifier. The plots highlight detection performance on abnormal points. Among abnormal data points, correctly classified points are marked with cyan, and misclassified data points are marked red. Black points are normal data points. While plots show that both the initial and modified models show high detection accuracy, the latter performs better.}
\label{fig:comparison_plot}
\end{figure}


\subsubsection{Improvement in the Modified Models}
In addition to the initial models, modified models which utilize the prediction result from the preceding time-step $\hat{y}_{t-1}$ are tested and the results are summarized in Table \ref{tbl:improvement_in_detection_performance_modified}. The table shows that they result in significant improvements in detection performance, almost reaching perfect detection performance. The datasets used in the experiment were especially suitable to show the benefit of this approach because normal and abnormal data occur consecutively in intervals. We visualize the detection performance of two models in Figure \ref{fig:comparison_plot}. Their difference is the usage of $\hat{y}_{t-1}$ to predict the current time-step. 
Moreover, similar to individual training, the models which utilize $\hat{y}_{t-1}$ work well in joint training settings with different sequence lengths in SFCs as in Table \ref{tbl:improvement_in_detection_performance_modified}.

\section{Conclusion}
Based on the sequential deep learning architectures like Uni-RNN, Bi-RNN, and Transformer, we proposed anomaly detection models composed of several neural network layers: feature mapping, encoder, readout, and classifier. The proposed models can learn sequential patterns over the sequence of VNFs and time steps in anomaly detection data, and they are compatible with the SFCs that have a varying number of VNF instances. The experiment results show improvement of F1-measure compared to the baselines. Furthermore, the results of joint training ensure that our proposed models could learn from different scenarios simultaneously. The future direction can be to study further the performance of the proposed models for different network traffic scenarios such as that of service for mobile connections.

\section*{Acknowledgement}
This research was supported by the Institute for Information \& communications Technology Promotion(IITP) grant funded by the Korea government(MSIT) (No. 2018-0-00749, Development of virtual network management technology based on artificial intelligence).

%
%
\bibliographystyle{IEEEtran}
\bibliography{bib2019}
\end{document}